\documentclass[10pt,conference,letterpaper]{IEEEtran}

\pdfoutput=1
\usepackage{amsmath,epsfig}
\usepackage{amsfonts}
\usepackage{times}
\usepackage{amssymb}
\usepackage{graphicx,subfig}
\usepackage{euscript}
\usepackage{algorithm, algorithmic}
\usepackage{color}
\usepackage{multirow}
\usepackage{booktabs}

\newcommand{\ru}    {\rule{0mm}{4mm}}

\newcommand{\be}    {\begin{equation}}
\newcommand{\ee}    {\end{equation}}

\newcommand{{\hx}}  {\widehat{x}}

\title{Image forgery detection based on the fusion of machine learning and block-matching methods}
\author{
{Davide Cozzolino, Diego Gragnaniello, Luisa Verdoliva}
\vspace{1.6mm} \\
\fontsize{10}{10}\selectfont\itshape
DIETI - University Federico II of Naples \\
Via Claudio 21, Naples - ITALY \\
\fontsize{9}{9}\selectfont\ttfamily\upshape
\{davide.cozzolino, diego.gragnaniello, verdoliv\}@unina.it
}

\begin{document}

\maketitle

\begin{abstract}
Dense local descriptors and machine learning have been used with success in several applications,
like classification of textures, steganalysis, and forgery detection.
We develop a new image forgery detector building upon some descriptors recently proposed in the steganalysis field
suitably merging some of such descriptors, and optimizing a SVM classifier on the available training set.
Despite the very good performance,
very small forgeries are hardly ever detected because they contribute very little to the descriptors.
Therefore we also develop a simple, but extremely specific, copy-move detector based on region matching
and fuse decisions so as to reduce the missing detection rate.
Overall results appear to be extremely encouraging.
\end{abstract}

\IEEEpeerreviewmaketitle

\section{Introduction}
This paper describes the strategy followed by the the GRIP team of the University Federico II of Naples (Italy)
to tackle phase 1 of the first IEEE IFS-TC Image Forensics Challenge on image forgery detection.

This team has been working in recent years on the forgery detection problem,
focusing on techniques based on camera sensor noise, a.k.a. PRNU (photo response non-uniformity) noise \cite{CPPSV_wmfsi.10,CPPSV_dsp.11,CPSV_mmsp.13,CPSV_TIFS.13}
and on techniques based on dense local descriptors and machine learning \cite{GPSV_bioms.13}.
Therefore, we decided to follow both these approaches for detection, on two separate lines of development,
with the aim of fusing decisions at some later time of the process.
Indeed, it is well known \cite{DGSV_iciap.13} that, given the different types of forgery encountered in practice,
and the wide availability of powerful photo-editing tools,
several detection approaches should be used at the same time and judiciously merged in order to obtain the best possible performance.
Based on this consideration,
we also followed a third line of development working on a technique for copy-move forgery detection which,
although applicable only to a fraction of the image set, provides very reliable results.

Unfortunately
it was very soon clear that the PRNU-based approach was bound to be of little use.
Lacking any information on the cameras used to take the photos,
we had to cluster the images based on their noise residuals
and estimate each camera's PRNU based on the clustered images.
However, more than 20\% of the test images could not be clustered at all
and in some cases the number of images collected in a cluster was too small to obtain a reliable estimate of the PRNU.

On the contrary,
techniques based on dense local descriptors appeared from the beginning very promising,
and we pursued actively this line of development, drawing also from the relevant literature in the steganalysis field.
Complementing such techniques with a simple copy-move detector, tuned so as to guarantee very high specificity,
lead us eventually to obtain very promising results.

The rest of the paper comprises only two sections,
one dealing with dense local descriptors and machine learning and the other with copy-move detection.
In each Section we provide experimental results obtained on the training set.

\section{Dense local descriptors for splicing detection}

Several techniques have been proposed in the last decade for splicing detection based on machine learning.
Major efforts have been devoted to find good statistical models for natural images
in order to single out the features that guarantee the highest discriminative power.
Often, in order to capture more meaningful statistics, transform-domain features have been used,
as in \cite{SCX_iwdw.07} where the image undergoes block-wise discrete cosine transform (DCT) with various block sizes
and first-order (histogram based) and higher-order (transition probabilities) features are collected and merged.
Given the good results obtained in terms of detection accuracy,
an expanded Markov-based scheme in DCT and DWT domains is followed in \cite{HLSH_PR.12}.
Interestingly,
the method proposed \cite{SCX_iwdw.07} was inspired by prior work carried out in steganalysis
which, despite the obvious differences with respect to the forgery detection field,
pursues a very similar goal, that is, detecting seemingly invisible alterations of the natural characteristics of an image.

The same path is followed in the forgery detection technique proposed in \cite{WDT_icip.09},
based on an approach proposed for steganalysis in \cite{ZSSX_icme.06,PBF_TIFS.10}.
The major contribution consists in deriving the features based on some co-occurrence matrices
computed on the thresholded prediction-error image (also called {\em residual image}).
In fact,
modeling the residuals rather than the pixel values is very sensible in these low-level methods (not based on image semantic),
since the image content does not help detecting local alterations and should be suppressed altogether.
In the context of forgery detection, in particular,
considering that splicing typically introduces sharp edges,
it is reasonable to characterize statistically some {\em edge image},
which can also be the output of a simple high-pass filter (like a derivative of first order).
As a further advantage,
the residual image has a much narrower dynamic range than the original one,
allowing for a compact and robust statistical description by means of co-occurrences.

The processing path outlined above, already proposed in \cite{ZSSX_icme.06}, can be therefore summarized in the following steps
\begin{enumerate}
\item   computation of the high-pass residuals;
\item   truncation and quantization;
\item   feature extraction based on co-occurrence matrices of selected neighbors;
\item   design of a suitable classifier on the training set.
\end{enumerate}
Given its compelling rationale, and some promising results obtained in the literature,
we will follow this path, here.
Nonetheless,
a large number of design choices must be made, beginning from the high-pass filter, to end with the classifier,
which impact heavily on the performance and require lengthy development and testing.
Fortunately,
we can rely on the precious results described in a recent work on steganalysis \cite{FK_TIFS.12},
where a large number of models have been considered and analyzed, and made available online to the research community \cite{Fridrich_web}.
Specifically,
in \cite{FK_TIFS.12} a number of different high-pass filters have been considered, both linear and nonlinear, with various supports,
different quantization and truncation strategies for the residues have been implemented and,
based on some preliminary experiments, the use of some selected groups of neighbors for co-occurrence computation has been suggested.
There is no doubt, as the Authors themselves point out,
that better design choices are possible, especially when aiming at slightly different goals,
but the wealth of models they provide allow for the rapid development and optimization of a specific processing chain,
which can be then improved, in part already in this work, under some specific respects.

\subsection{Implemented method}

In \cite{FK_TIFS.12} 39 different high-pass filters are proposed,
which work on the grayscale version of the original image obtained by standard conversion.
All such filters are extremely simple,
since their goal is to highlight minor variations w.r.t. to typical behaviors.
Typical example are the first order horizontal linear and symmetric nonlinear filters defined by
\begin{eqnarray}
    r_{i,j} & = & x_{i,j+1}-x_{i,j} \nonumber \\
    r_{i,j} & = & \min [ (x_{i,j+1}-x_{i,j}) , (x_{i+1,j}-x_{i,j}) ] \nonumber
\end{eqnarray}
Fig.1 shows the effect of applying one of such filters to a training image of the challenge.
Of course, it is not obvious by visual inspection that the forged region (in black in the ground-truth)
exhibits characteristics different from those of the rest of the image,
and such to allow the identification of the forgery.

\setlength\fboxsep{0.5pt}\setlength\fboxrule{0.2pt}

\begin{figure}[t]
\centering
\includegraphics[width=0.15\textwidth]{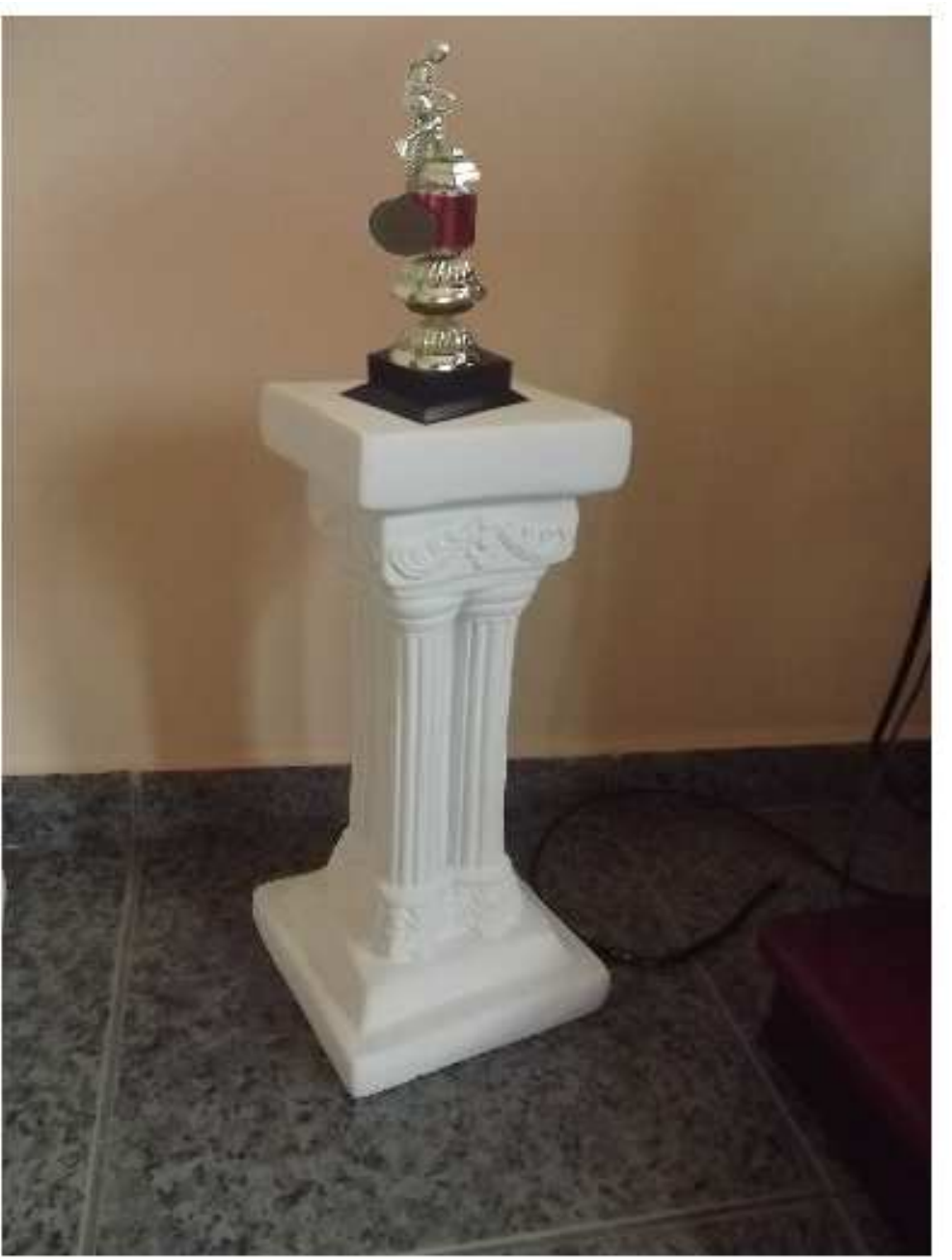}
\fbox{\includegraphics[width=0.15\textwidth]{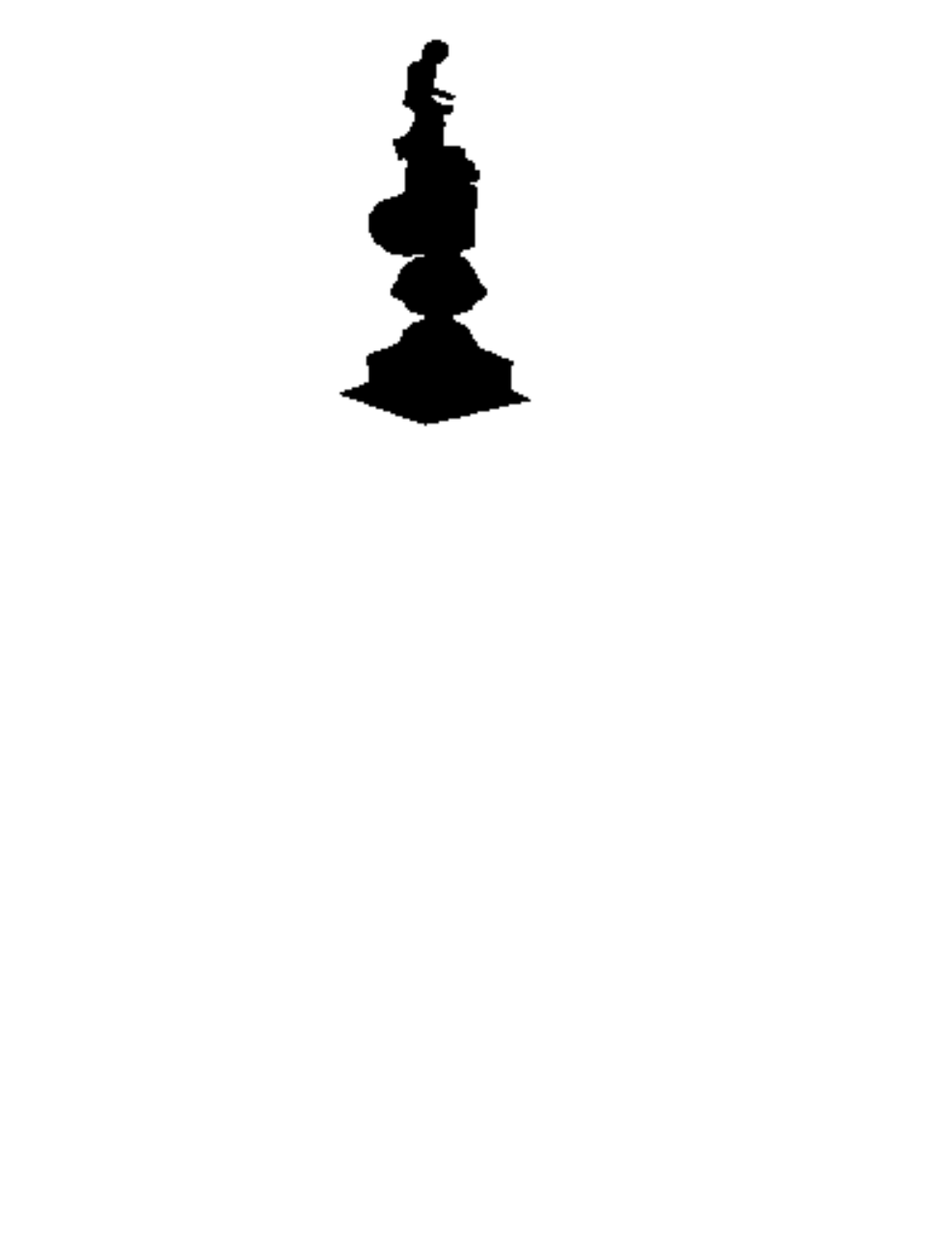}}
\includegraphics[width=0.15\textwidth]{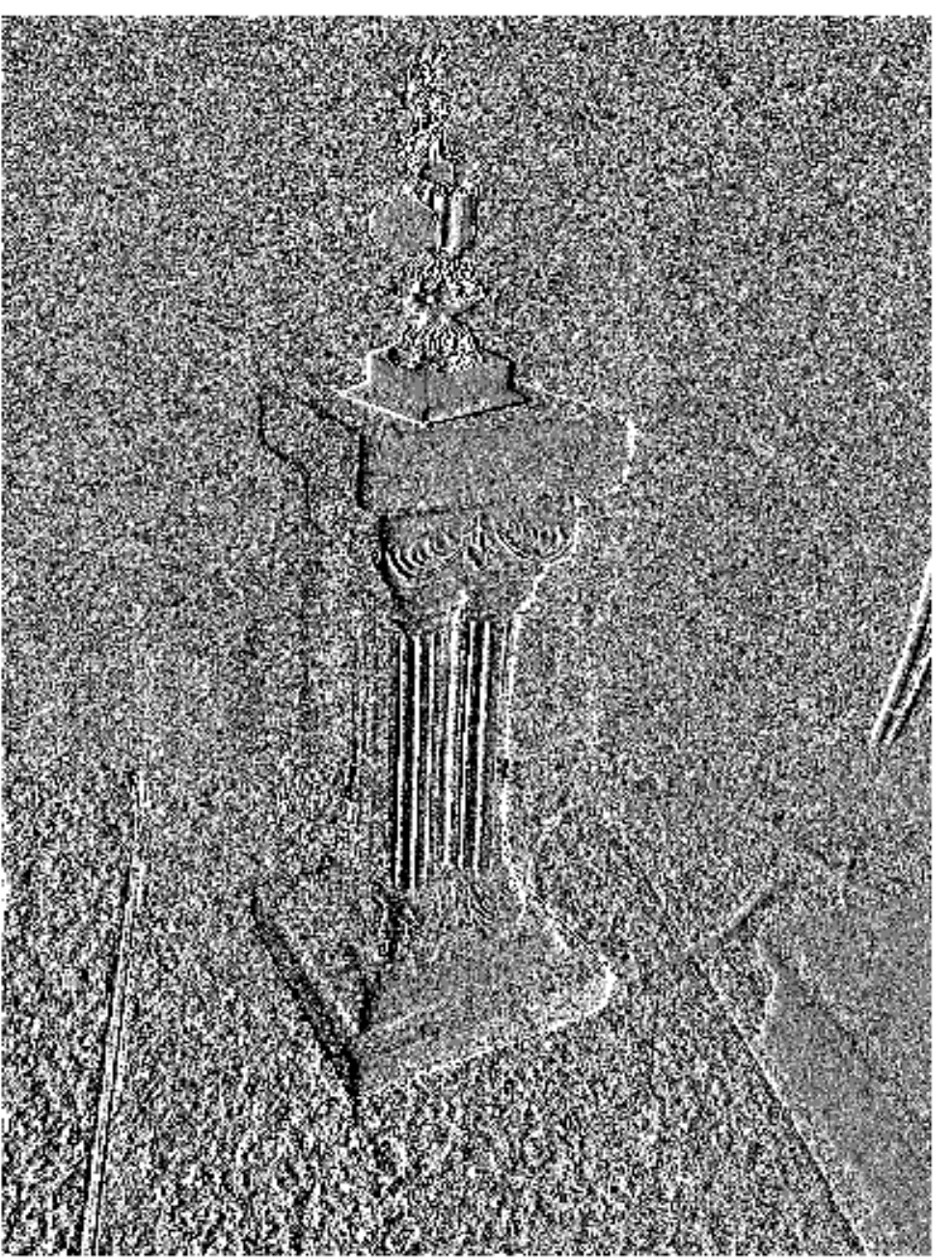}
\caption{A training image with its ground truth and an example residual image.}
\label{fig:residue}
\end{figure}

Residuals are in general real-valued and, although typically small, are defined on a wide range.
To enable their meaningful characterization in terms of co-occurrence they must be quantized and truncated.
Following \cite{FK_TIFS.12} we use
\[
    \widehat{r}_{ij} = {\rm trunc}_T({\rm round}(r_{ij}/q))
\]
with $q$ the quantization step and $T$ the truncation value.
We keep using $T$=2 to limit the matrix size but consider exclusively $q$=1,
partly to reduce complexity, but mainly to limit the risk of overfitting to our training set.
Each quantized residual can eventually take on 5 values, from -2 to +2.
We then compute co-occurrences on four consecutive pixels along the same row or column,
obtaining 625 entries, which can be highly reduced thanks to symmetries.

In the classification phase we depart significantly from the reference technique,
due to the overfitting problem mentioned before.
In fact, each individual model comprises 169 features for linear filters
and 325 for non linear ones,
a number large but still adequate for a training set comprising about 1500 images (450 fake and 1050 pristine), as in our case.
Merging all models, however,
would lead to a much larger number of features probably too large to expect a meaningful training.
The Authors of \cite{FK_TIFS.12} dealt successfully with this problem using a suitable {\em ensemble} classifier \cite{FK_TIFS.12b}.
In this challenge, however, we have a training set about ten times smaller,
which raises serious doubts on the chances of success of this approach.

We decided therefore to test each model individually,
relying heavily on cross validation to gain a reasonable insight into their actual performance.
In each experiment, we selected at random 5/6 of the pristine images and 5/6 of the fake ones to train a SVM classifier.
The remaining images of each class were then used to test the trained classifier.
To reduce randomness, each experiment was repeated 18 times, selecting the training and test set at random, and results were eventually averaged.
Fig.2(top) shown the results for the 39 models considered, in terms of expected score, defined as
\newcommand{\wF}{\widehat{F}}
\newcommand{\wP}{\widehat{P}}
\[
    S = \frac{\Pr(\wF|F)+\Pr(\wP|P)}{2}
\]
with $P[F]$ indicating the event ``image pristine[fake]'' and $\wP[\wF]$ the event ``decision pristine[fake]'', respectively.
For several models the predicted score is in the order of 94\%, hence very promising.
Then we tried to merge the features of a limited number of models, up to four, not to exceed the number of training images.
Results are reported in Tab.I in terms of score obtained before and after merging.
They show a limited improvement, if any, over the best single-model classifier, and a non-monotonic behavior, ringing an alarm bell on stability.

To improve robustness, we considered a different measure of performance.
For each SVM classifier, we moved the separating hyperplane along the orthogonal direction, and built the corresponding ROC.
Then we computed, for each model, the Area Under the receiver operating Curve (AUC),
because a large AUC implies not only a good performance in the best operating point, but also robustness w.r.t. changing conditions.
Fig.2(bottom) shows results.
We then tried merging the best models selected with this criterion, obtaining the results reported in Tab.II.
This time, performance improves monotonically, supporting the use of a merged set of features selected with this latter choice.

Eventually,
our SVM classifier uses the merging of all the features of models 17 31 34 and 36, and is trained over the whole phase-1 training set.

\begin{figure}
\centering
{\includegraphics[width=0.92\columnwidth]{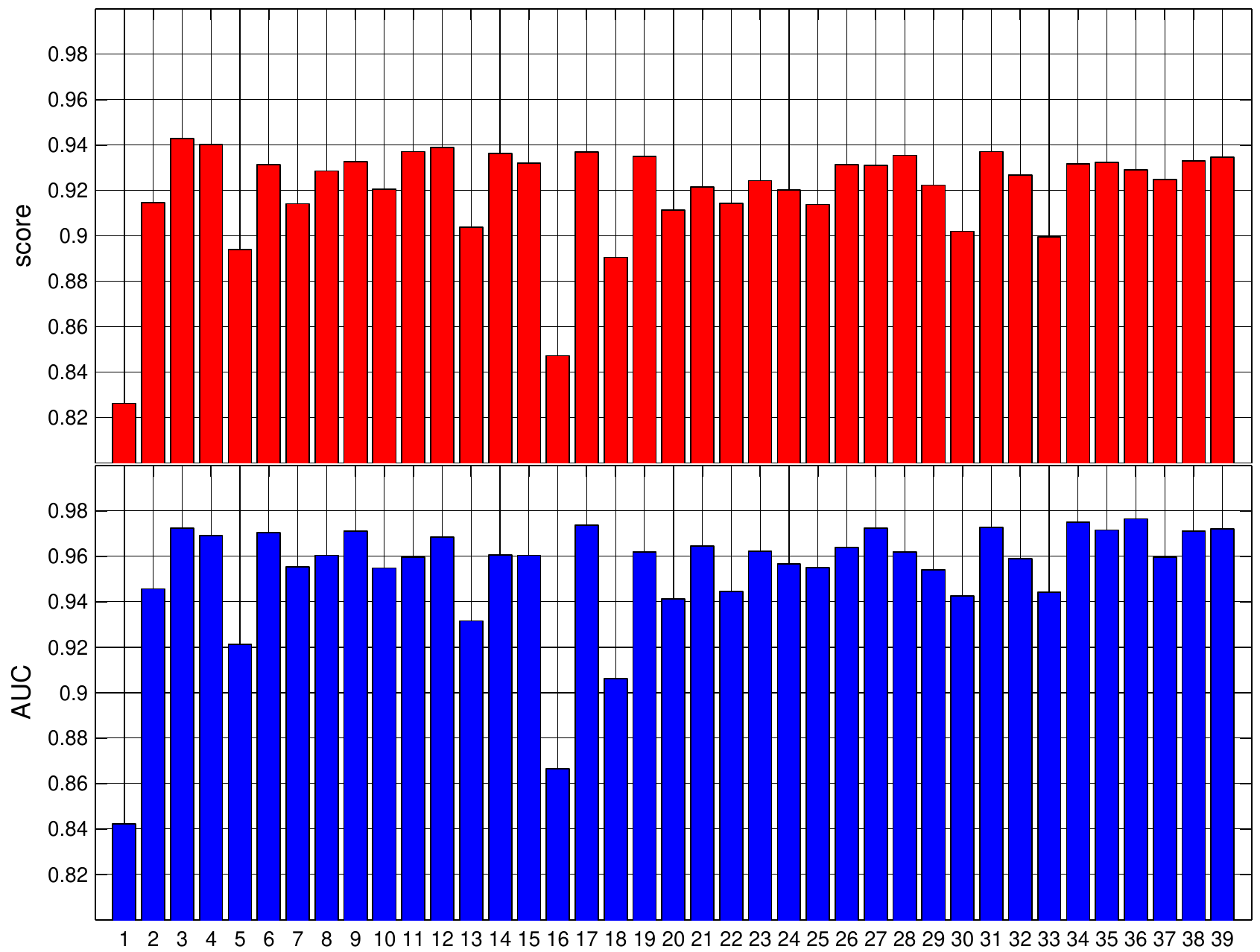}}
\caption{Scores (top) and AUC (bottom) for all models.}
\label{fig:bars}
\end{figure}

\begin{table}[!htb]
\centering
\begin{tabular}{|c|c|c|c|c|} \hline
\ru Model     & Type                    &  Score      &  AUC        & Score/merg.   \\ \hline
\ru 3         & non linear, 1st order   &  0.9429     &  0.9724     & 0.9429        \\ \hline
\ru 4         & non linear, 1st order   &  0.9403     &  0.9693     & 0.9154        \\ \hline
\ru 12        & non linear, 2nd order   &  0.9389     &  0.9685     & 0.9415        \\ \hline
\ru 11        & non linear, 2nd order   &  0.9371     &  0.9595     & 0.9163        \\ \hline
\end{tabular}
\caption{Score obtained before and after merging by the top-score individual models.}
\label{top-score}
\end{table}

\begin{table}[!htb]
\centering
\begin{tabular}{|c|l|c|c|c|} \hline
\ru  Model    & Type                  &  Score      &  AUC        & Score/merg.   \\ \hline
\ru  36       & linear, 3rd order     &  0.9289     &  0.9765     & 0.9289        \\ \hline
\ru  34       & linear, 1st order     &  0.9316     &  0.9751     & 0.9462        \\ \hline
\ru  17       & non linear, 3rd order &  0.9369     &  0.9736     & 0.9481        \\ \hline
\ru  31       & non linear, square 5$\times$ 5      &  0.9371     &  0.9727     & 0.9531        \\ \hline
\end{tabular}
\caption{Score obtained before and after merging by the top-AUC individual models.}
\label{top-AUC}
\end{table}

\section{Copy-move detection by PatchMatch}

Many algorithms have been proposed in the literature for copy-move forgery detection,
typically based on matching techniques, {\it e.g.}, \cite{LG_crv.06,PL_TIFS.10,DYMT_FSI.13}.
The major source of difference between them resides in the hypotheses made on the nature of the forgery.
In particular,
detection performance and algorithm complexity depend heavily
on the size of the copied region,
on its content as compared with the target region background, and
on the presence/absence of further processing on such regions, such as rotation, resizing, change of illumination, and so on.
Algorithms aiming at the detection of large copy-moves characterized by rigid translation can be quite simple,
while they grow more and more complex as constraints are relaxed including new potential targets.

Let us focus, for the time being, on the simplest possible problem,
in which one or more patches of the image are copied somewhere else by pure translation.
Then,
a pretty general detection algorithm might comprise the following steps
\begin{enumerate}
\item   computation of a dense motion-vector field;
\item   segmentation of the field in regions characterized by homogeneous motion vectors;
\item   elimination of candidate matching regions based on size, matching error, and other criteria.
\end{enumerate}
In the hypotheses cited above, and barring pathological cases such as uniformly dark or saturated areas,
any copy-move forgery of reasonable size can be detected easily, and with very high confidence.
Indeed,
it is very difficult to find {\em identical} regions in a pristine natural image,
a chance that becomes totally negligible as the region size grows larger.

If we abandon the strong constraints considered before, things become quickly much more difficult.
Rotation and resizing imply a non-constant motion field in copied areas,
and also an intensity mismatch due to pixel interpolation, further increased by possible changes of illumination.
Algorithms have been proposed to deal with all these problems but,
besides being more complex
they provide weaker guarantees on the absence of false alarms.
For example, in a highly textured areas, like a close up of a tree,
it might be very difficult to decide whether a certain leaf is a rotated and rescaled version of another or not.

These considerations
serve to justify some important design choices in the development and fine-tuning of our
approach.
Consider, in fact,
that we are trying to optimize
the performance of a composite detector
obtained through the suitable fusion with a machine-learning method.
Under this perspective,
the marginal accuracy of the copy-move detector becomes immaterial w.r.t. its contribution to the overall performance.
Preliminary experiments show that the detector described in the previous Section is characterized by an excellent and well balanced performance on the training set,
with very low missing-detection and false-alarm rates (e.g. 0.0726 and 0.0213, respectively, for the best score).
The copy-move detector cannot reduce the overall false-alarm rate,
since its ``pristine'' decision means only that there is (probably) no copy-move forgery, but a splicing could still be present.
However, it can help reducing the missing-detection rate,
by revealing all those copy-move forgeries that have escaped the previous detector,
very likely because they are too small to impact on the descriptor.
To this end, it is necessary that it be extremely specific, assuring that its ``fake'' decision is very reliable.
Based on these considerations,
we develop an algorithm aimed basically at detecting rigid-translation copy-move forgeries, with little tolerance for other forms of processing,
thus ensuring a very high specificity.

\subsection{Implemented method}

As outlined before,
our first processing step is the computation of a dense motion vector field based on block matching.
Carrying out an exact search for each block of the image, however, is exceedingly burdensome,
and in fact this step is often replaced by simpler, though less reliable methods, {\it e.g.} \cite{LG_crv.06}.
Here we resort to PatchMatch,
an iterative algorithm recently proposed for image editing applications \cite{BSFG_TG.09, Barnes_site}.
Patchmatch provides a very accurate and regular motion field,
but we chose it primarily for its rapid convergence, which makes it about 100 times faster than exact methods,
allowing us to process in reasonable time a large database of images.

We use 7$\times$7 pixel patches, a size that guarantees a good compromise among accuracy, resolution and speed.
All image pixels are preliminarily adjusted to unitary norm,
in order to single out copy-moves also in the presence of some intensity adjustments.
After computing the motion vector field,
we carry out a filtering on both components to identify regions with homogeneous motion.
Choosing an appropriate filter, we can also identify regions where motion vectors slowly increase or decrease linearly,
thus identifying also copy-moves with moderate resizing.

Once a relatively large region with uniform motion is identified,
all matches obtained in perfectly flat areas, as in presence of saturation, are removed;
in addition, very small regions are deleted automatically through morphological filtering.
Eventually, after elimination of unsuitable candidates, the image is classified as fake if at least one duplicated region is detected.
To find also rotated copy-moves,
we simply repeat the procedure for a number of rotations of the image, taking advantage of PatchMatch speed.
Our experiments showed that a sampling step of 15 degrees guarantees accurate detection.

Fig.3 shows three images with copy-move forgeries, the corresponding ground truth, and the detection map output by our method.
Note that the forgery is easily detected, and the map is quite accurate,
although the original and copied regions are not distinguished from one another.
Turning to results,
our method detects only 271 of the 450 fakes of the training set, most of the other cases being splicing.
However it declares fakes only 5 of the 1050 pristine images,
and therefore its specificity, 99.52\%, is extremely high as was desired from the beginning.
We exploit this property in the final decision, by declaring a fake when at least one of the methods does.
Consequently, the score on the training set increases from 0.9531 to 0.9737.
Note that using this strategy we obtained the best score of phase 1A of the Challenge
with 0.9429.

\setlength\fboxsep{0.8pt}\setlength\fboxrule{0.5pt}

\begin{figure}[t]
\centering
\includegraphics[width=0.15\textwidth]{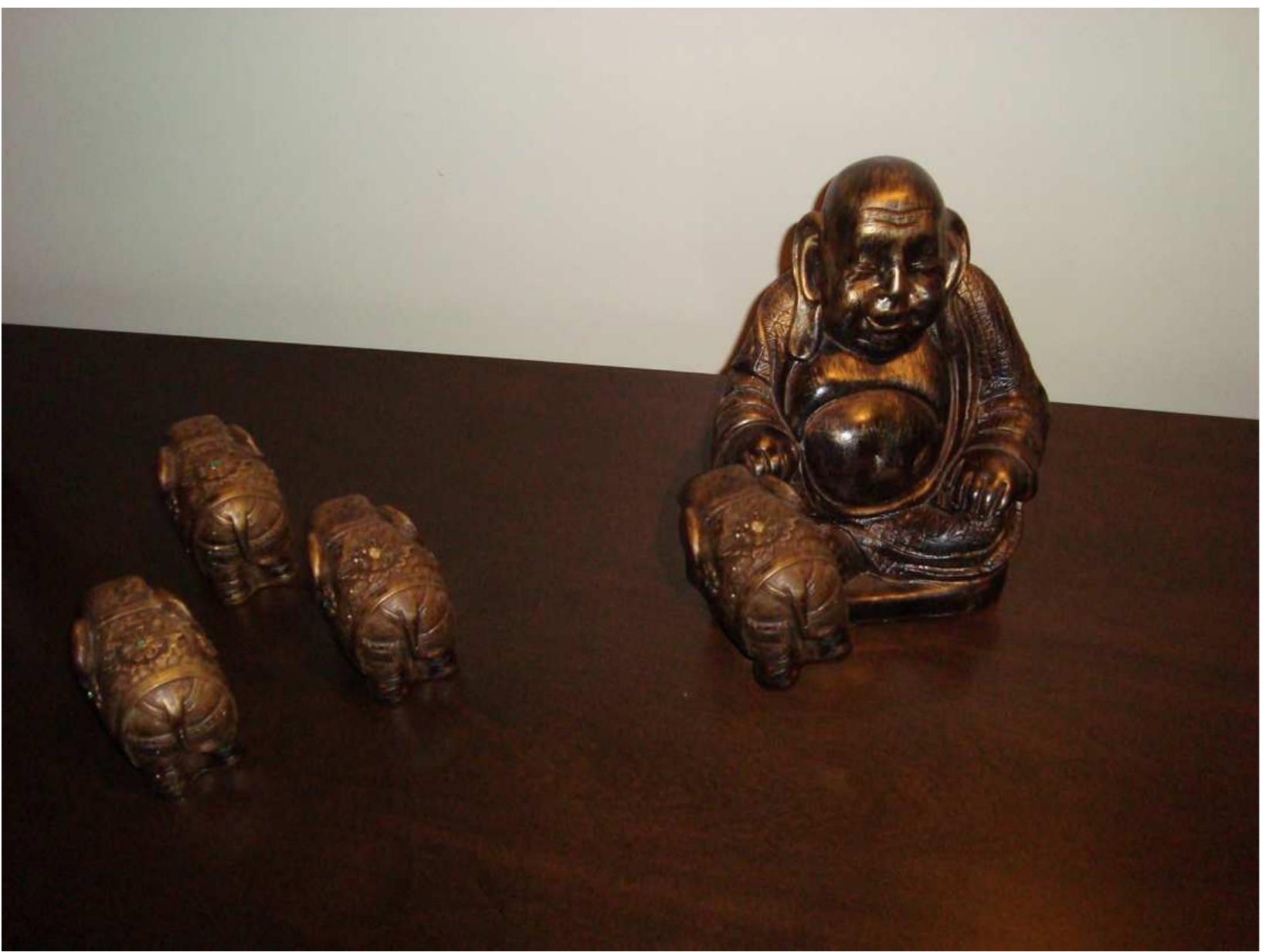}
\fbox{\includegraphics[width=0.15\textwidth]{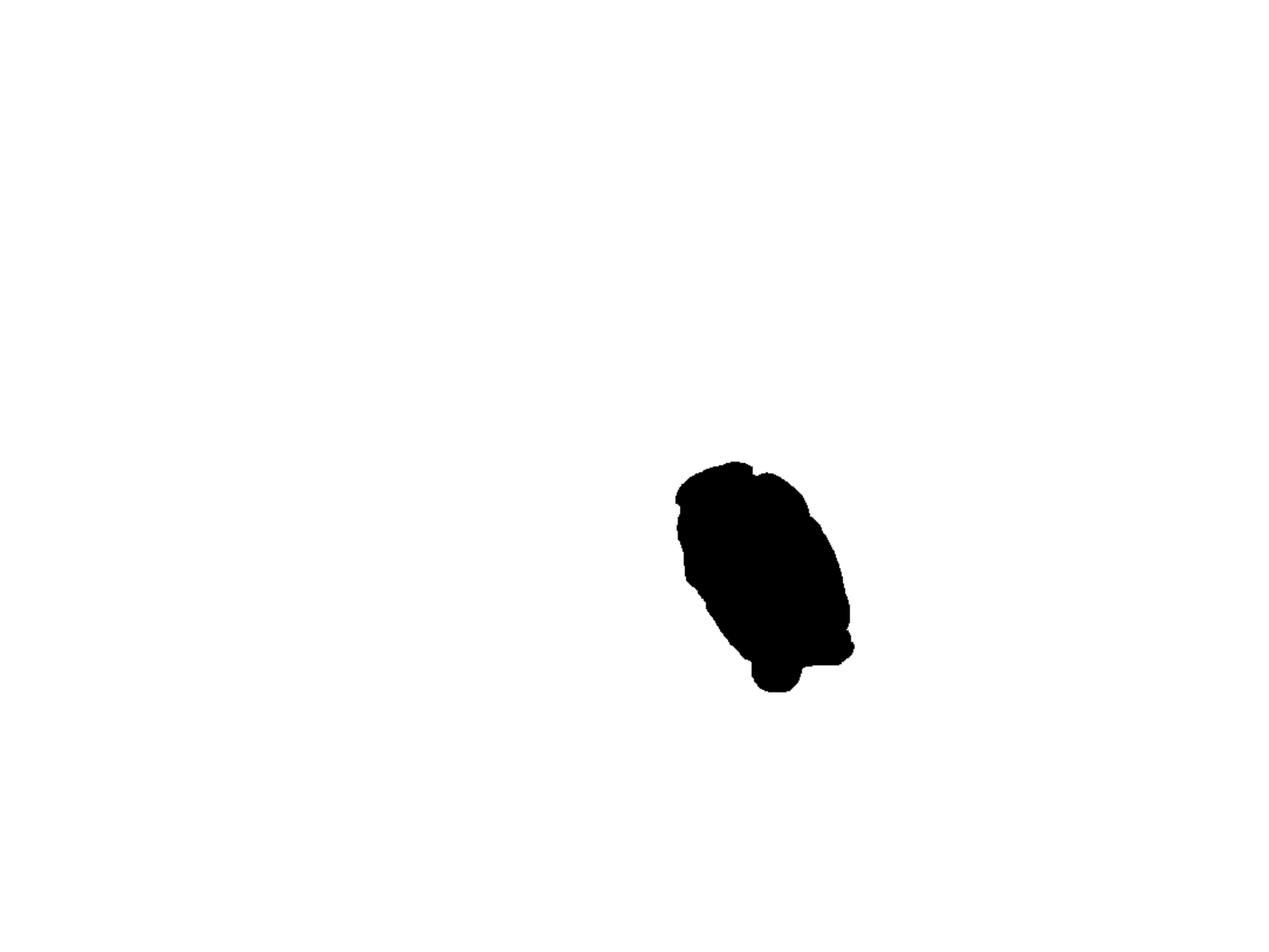}}
\fbox{\includegraphics[width=0.15\textwidth]{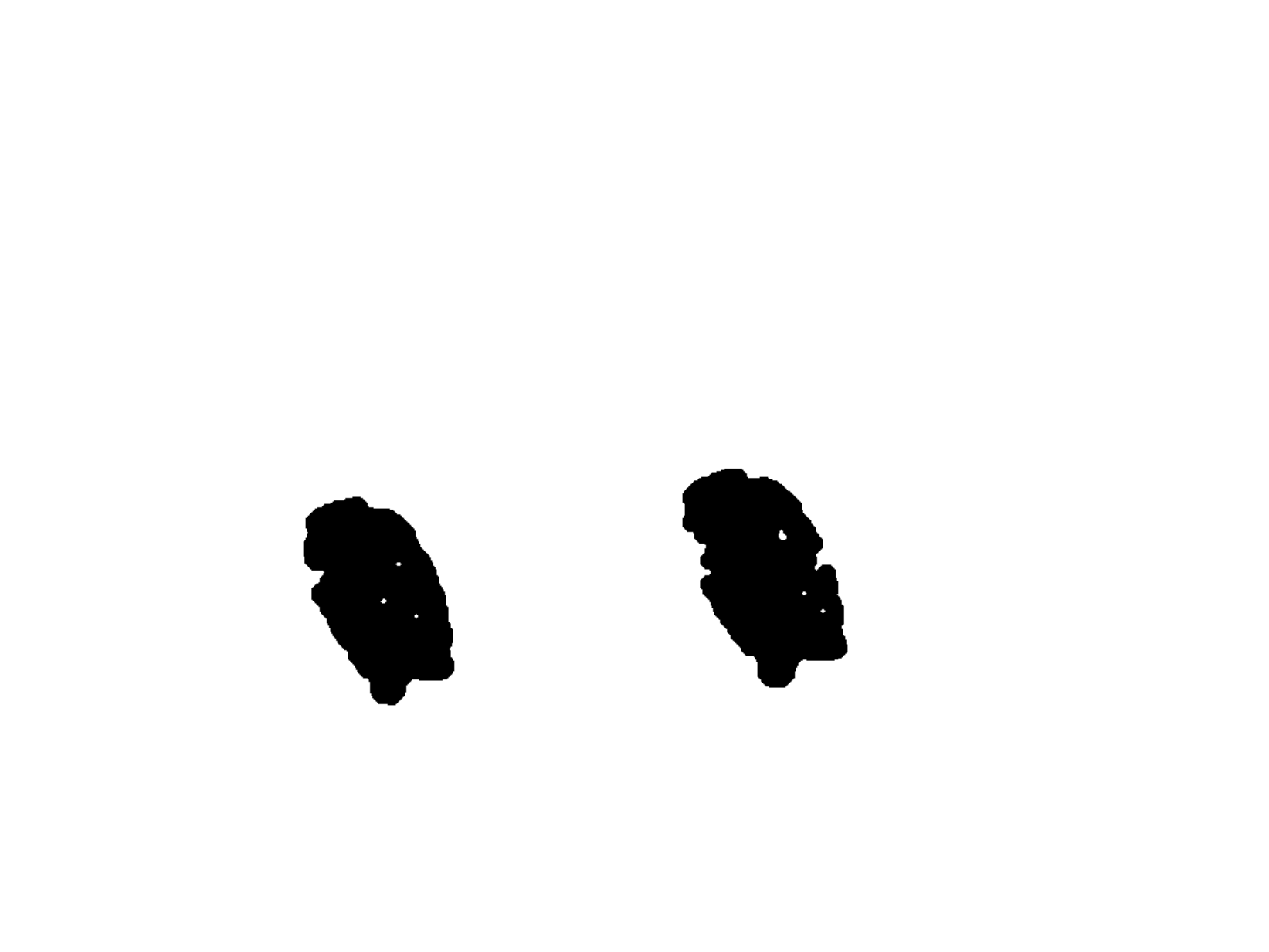}}

\vspace{2mm}
\includegraphics[width=0.15\textwidth]{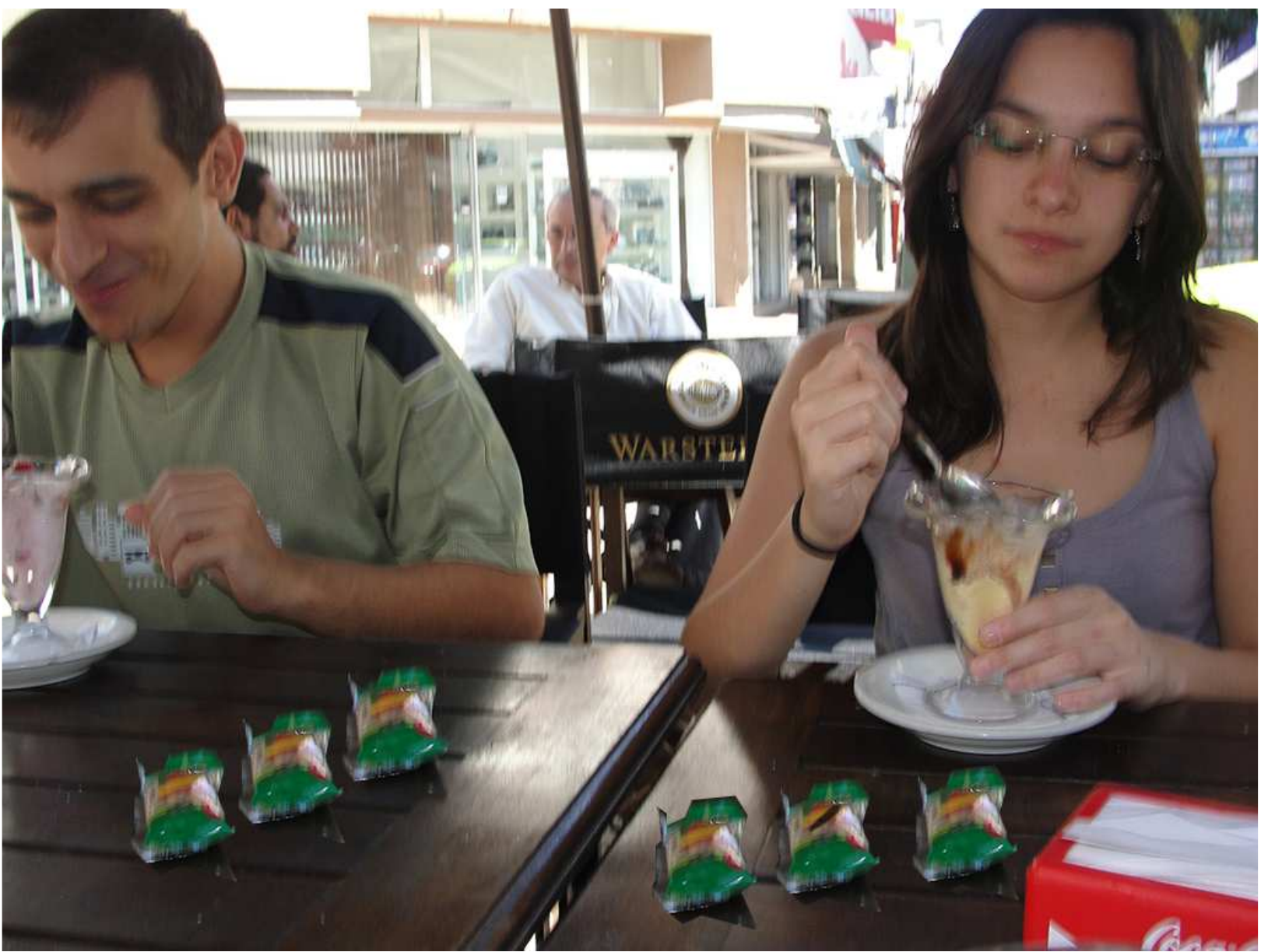}
\fbox{\includegraphics[width=0.15\textwidth]{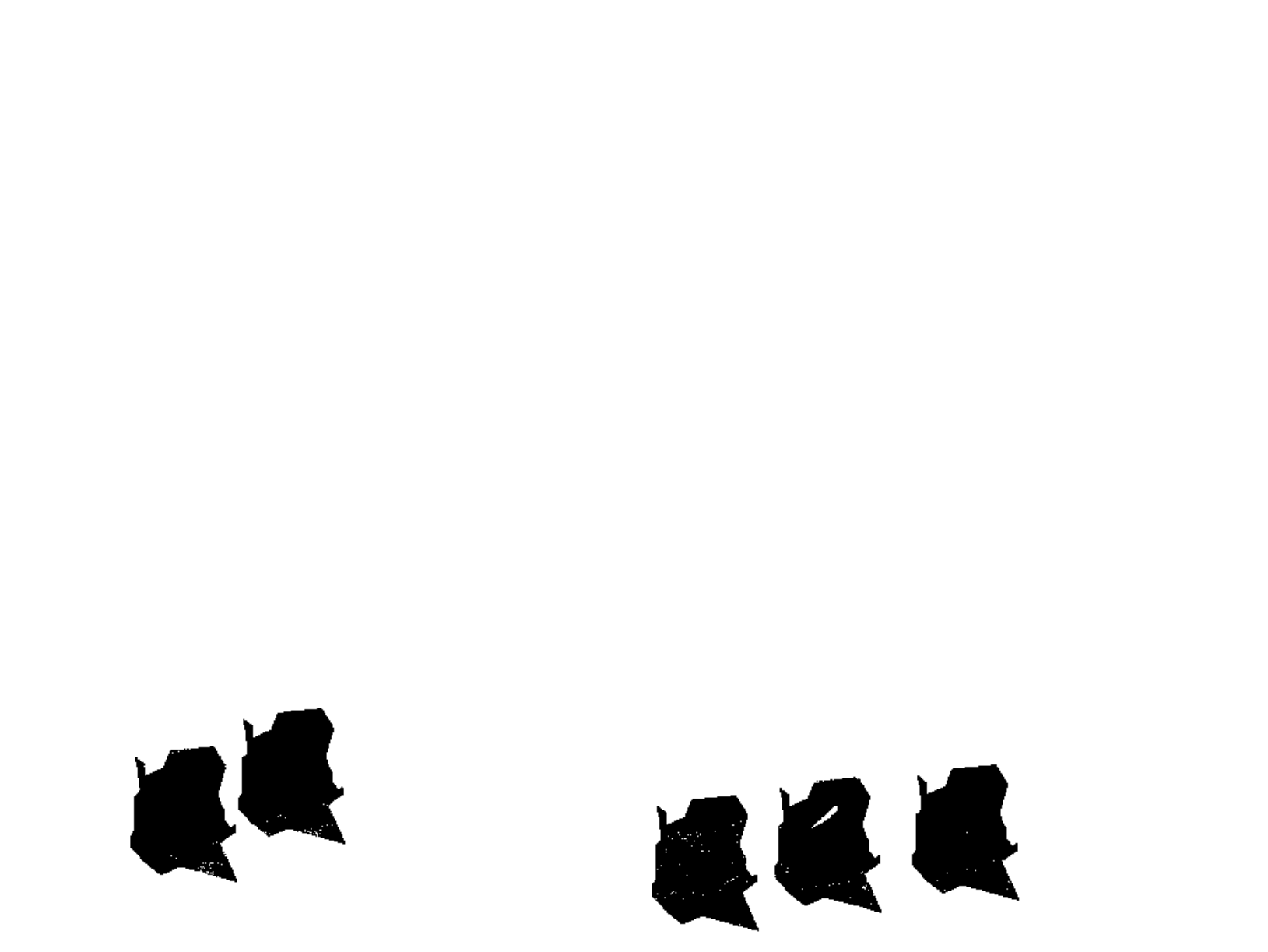}}
\fbox{\includegraphics[width=0.15\textwidth]{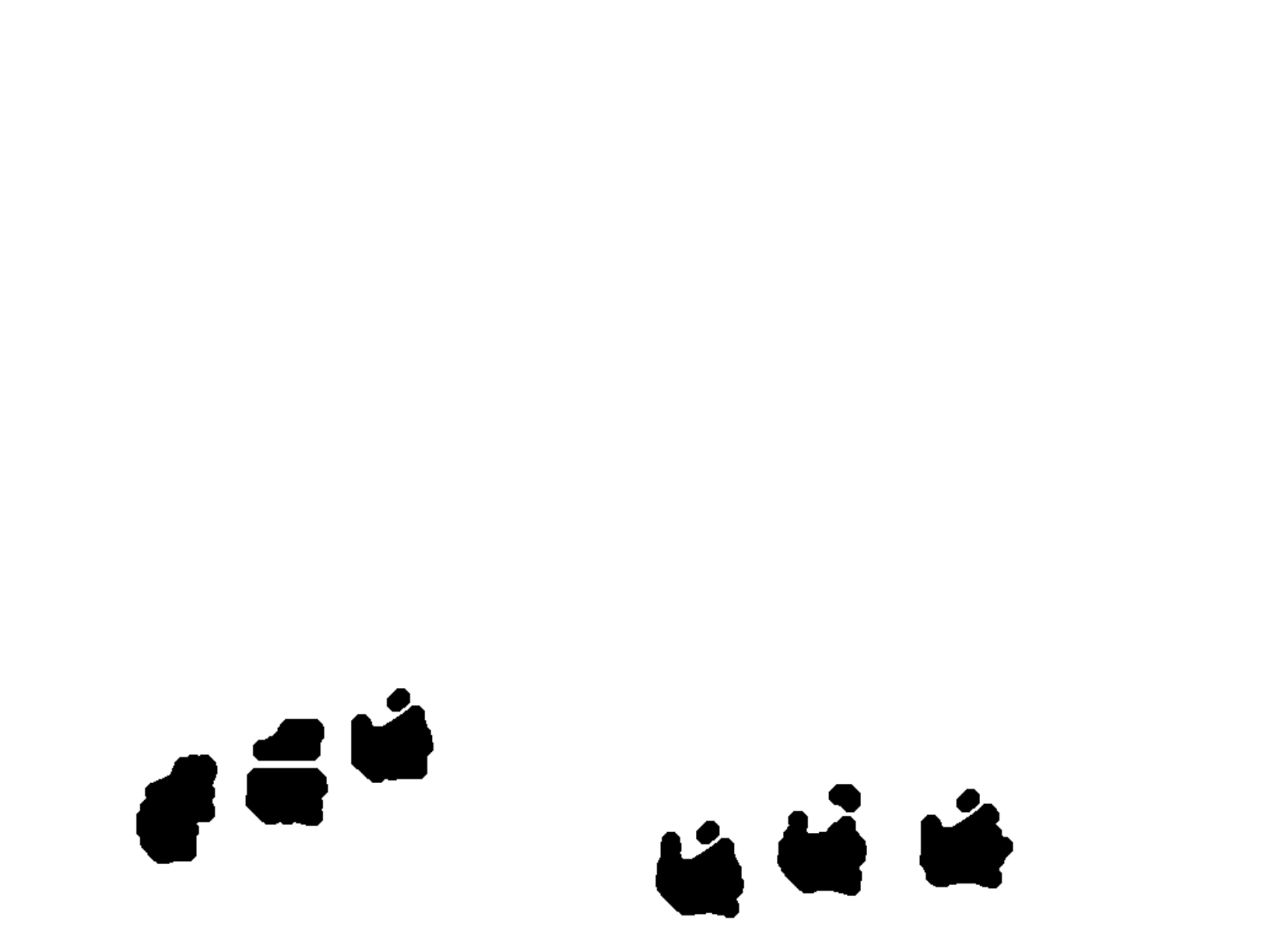}}

\vspace{2mm}
\includegraphics[width=0.15\textwidth]{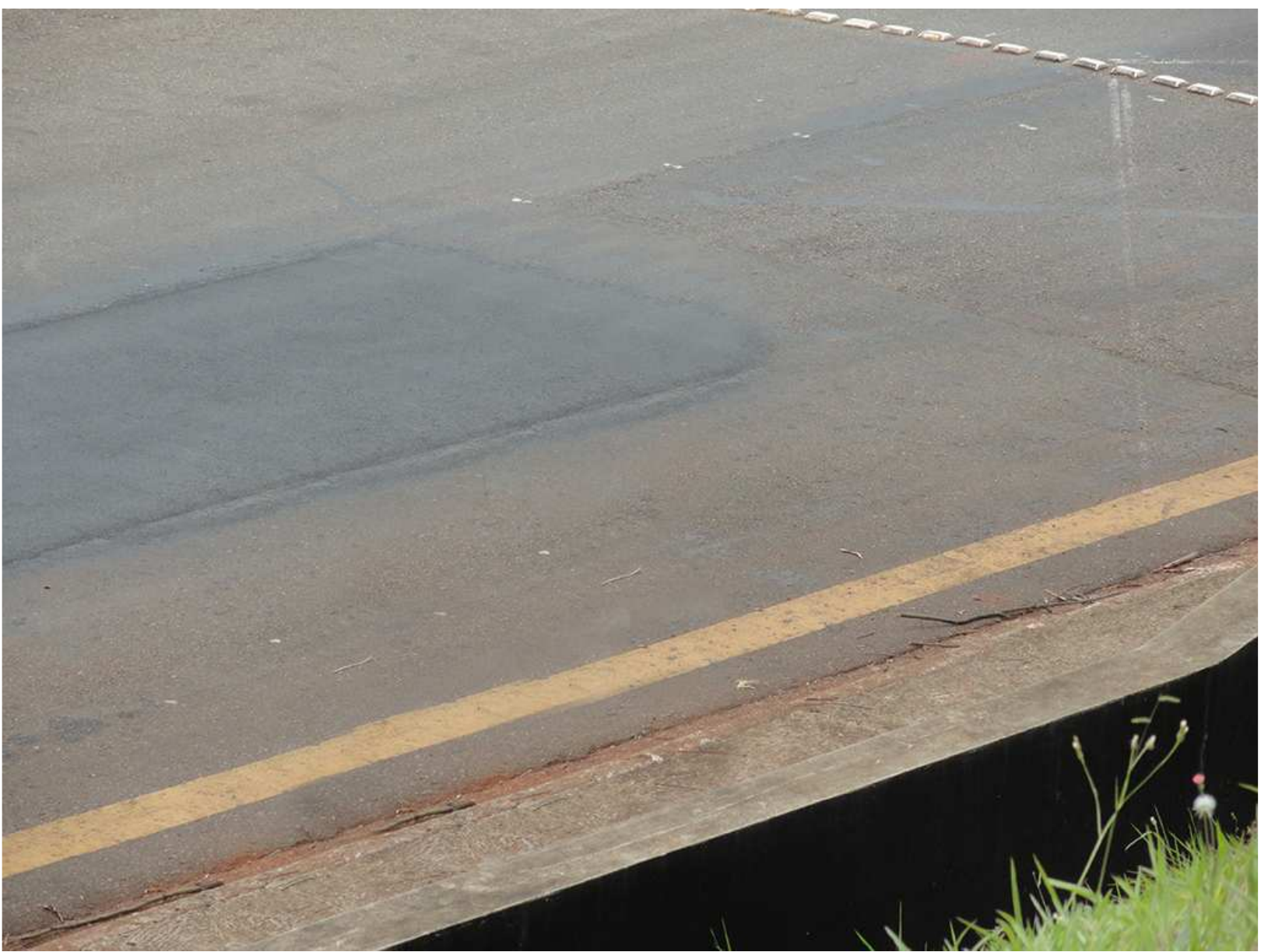}
\fbox{\includegraphics[width=0.15\textwidth]{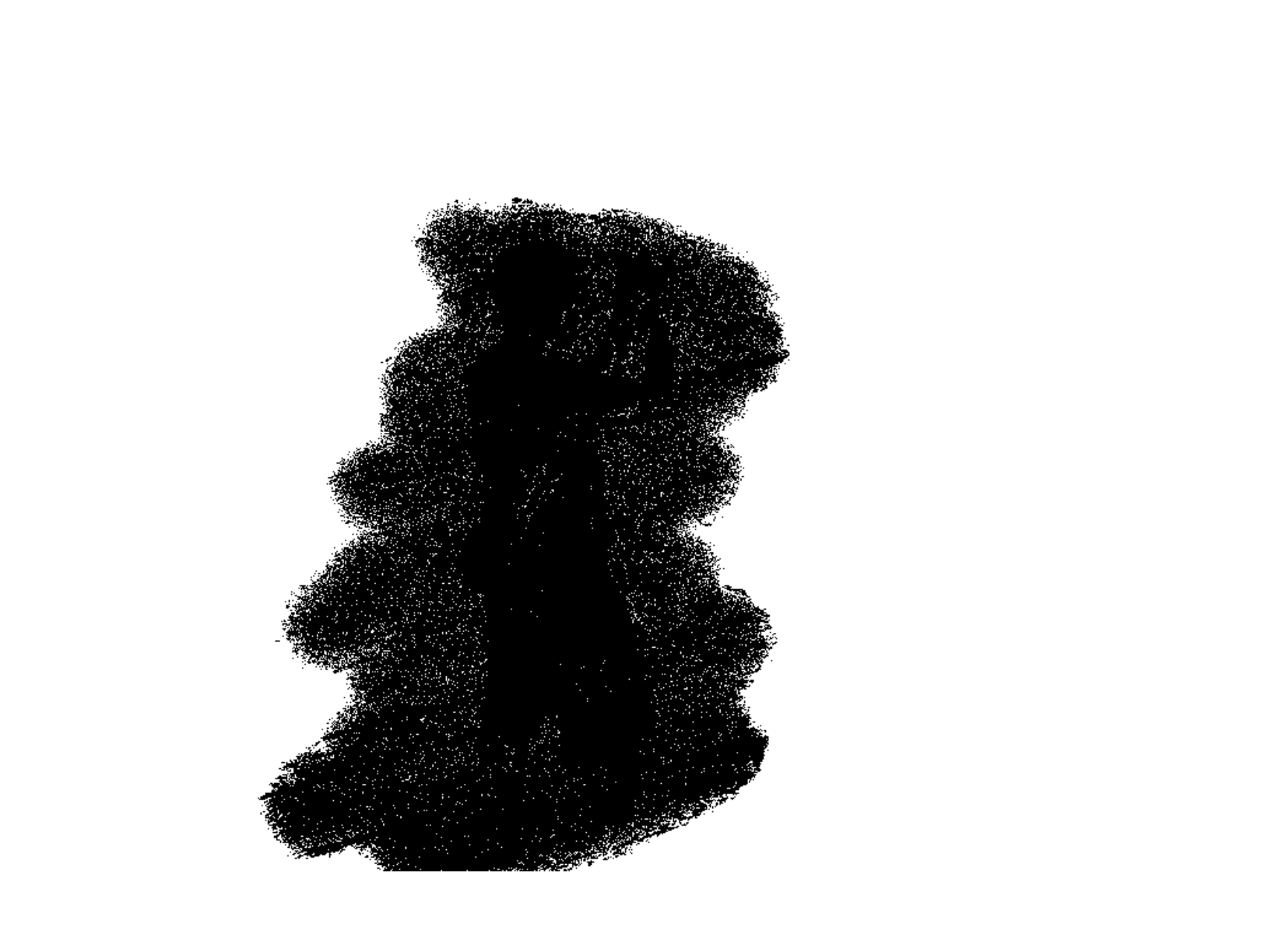}}
\fbox{\includegraphics[width=0.15\textwidth]{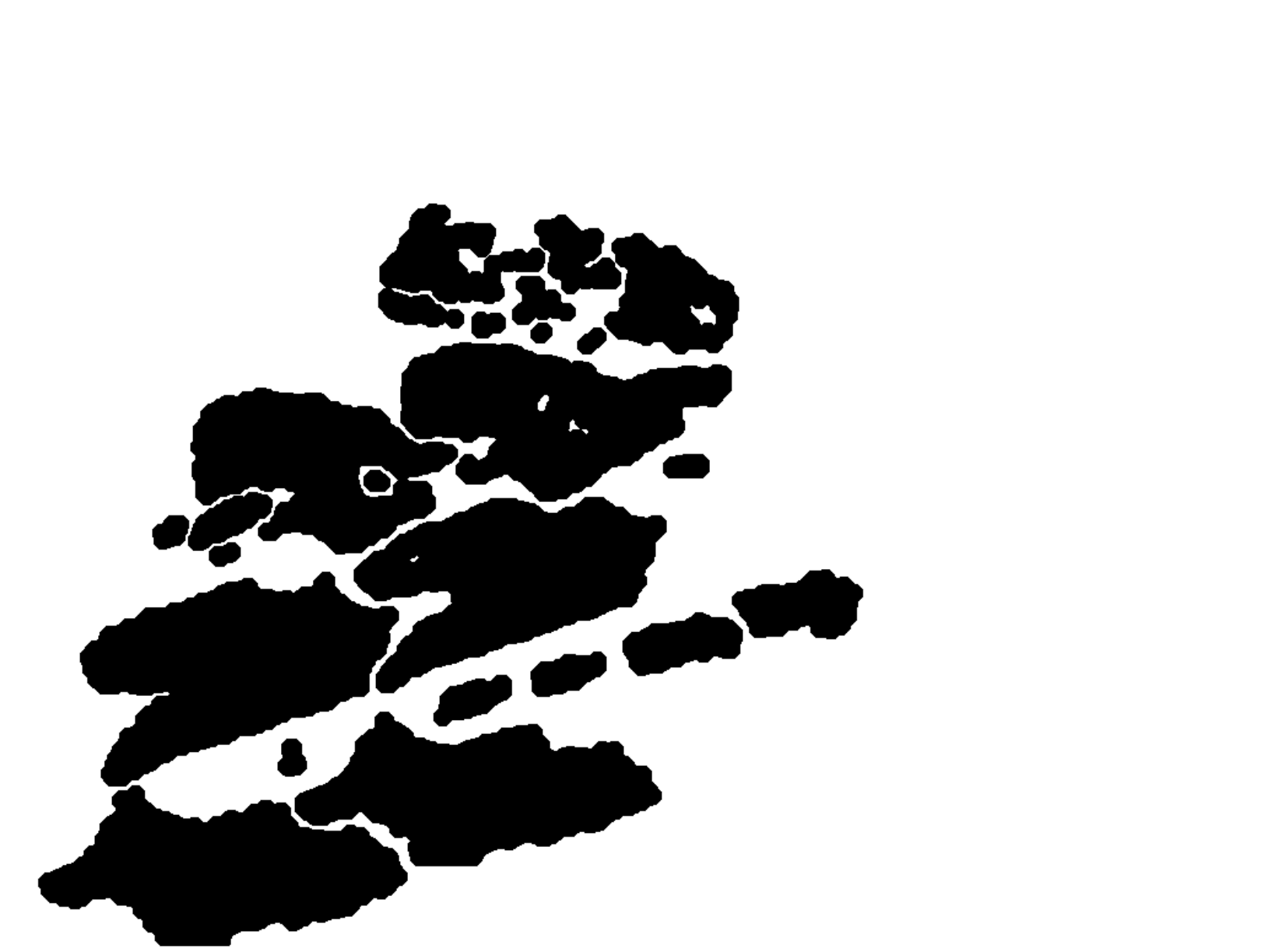}}
\caption{Three training images with copy-move forgeries, their ground truth, and detection maps output by our method.}
\label{fig:copymoves}
\end{figure}

\end{document}